\documentclass[letterpaper, 10pt, conference, twoside]{style/ieeeconf}    %

\makeatletter
\let\NAT@parse\undefined
\makeatother

\usepackage{dblfloatfix}

\usepackage{graphicx}
\usepackage{subcaption}
\usepackage{amsmath}
\usepackage{amssymb}
\usepackage{booktabs}
\usepackage[table,xcdraw]{xcolor}
\usepackage{float}
\usepackage{tabularx}
\usepackage{multirow}
\usepackage{siunitx}
\usepackage[font=small]{caption}
\usepackage{balance}
\usepackage{url}
\usepackage{amsmath}
\DeclareMathOperator*{\argmax}{arg\,max}

\usepackage{bm}

\usepackage{cite}

\usepackage{color}
\definecolor{mygreen}{RGB}{26, 148, 49}

\newcommand{\reffig}[1]{Fig.~\ref{#1}}

\newcommand{\refsec}[1]{Sec.~\ref{#1}}

\usepackage{float}
\usepackage[multiple]{footmisc}

\IEEEoverridecommandlockouts   

\begin{document}
\title{\LARGE \bf Perceptual Factors for Environmental Modeling \\ in Robotic Active Perception}

\author{David Morilla-Cabello, Jonas Westheider, Marija Popović, and Eduardo Montijano%
\thanks{This work was partially funded by the Deutsche Forschungsgemeinschaft (DFG, German Research Foundation) under Germany’s Excellence Strategy - EXC 2070 – 390732324. D. Morilla-Cabello and E. Montijano are with the Instituto de Investigaci\'on en Ingenier\'ia de Arag\'on, Universidad de Zaragoza, Spain.
J. Westheider and M. Popović are with the Institute of Geodesy and Geoinformation, University of Bonn, Germany.
\texttt{\small \{davidmc, emonti\}@unizar.es}}
}
\maketitle

\begin{abstract}

Accurately assessing the potential value of new sensor observations is a critical aspect of planning for active perception. This task is particularly challenging when reasoning about high-level scene understanding using measurements from vision-based neural networks. Due to appearance-based reasoning, the measurements are susceptible to several environmental effects such as the presence of occluders, variations in lighting conditions, and redundancy of information due to similarity in appearance between nearby viewpoints. To address this, we propose a new active perception framework incorporating an arbitrary number of perceptual effects in planning and fusion. Our method models the correlation with the environment by a set of general functions termed \textit{perceptual factors} to construct a perceptual map, which quantifies the aggregated influence of the environment on candidate viewpoints. This information is seamlessly incorporated into the planning and fusion processes by adjusting the uncertainty associated with measurements to weigh their contributions. We evaluate our perceptual maps in a simulated environment that reproduces environmental conditions common in robotics applications. Our results show that, by accounting for environmental effects within our perceptual maps, we improve in the state estimation by correctly selecting the viewpoints and considering the measurement noise correctly when affected by environmental factors. We furthermore deploy our approach on a ground robot to showcase its applicability for real-world active perception missions.
\end{abstract}

\begin{figure}[t]
    \centering
    \includegraphics[width=0.98\columnwidth, trim={0 0.5cm 0 0.5cm},clip]{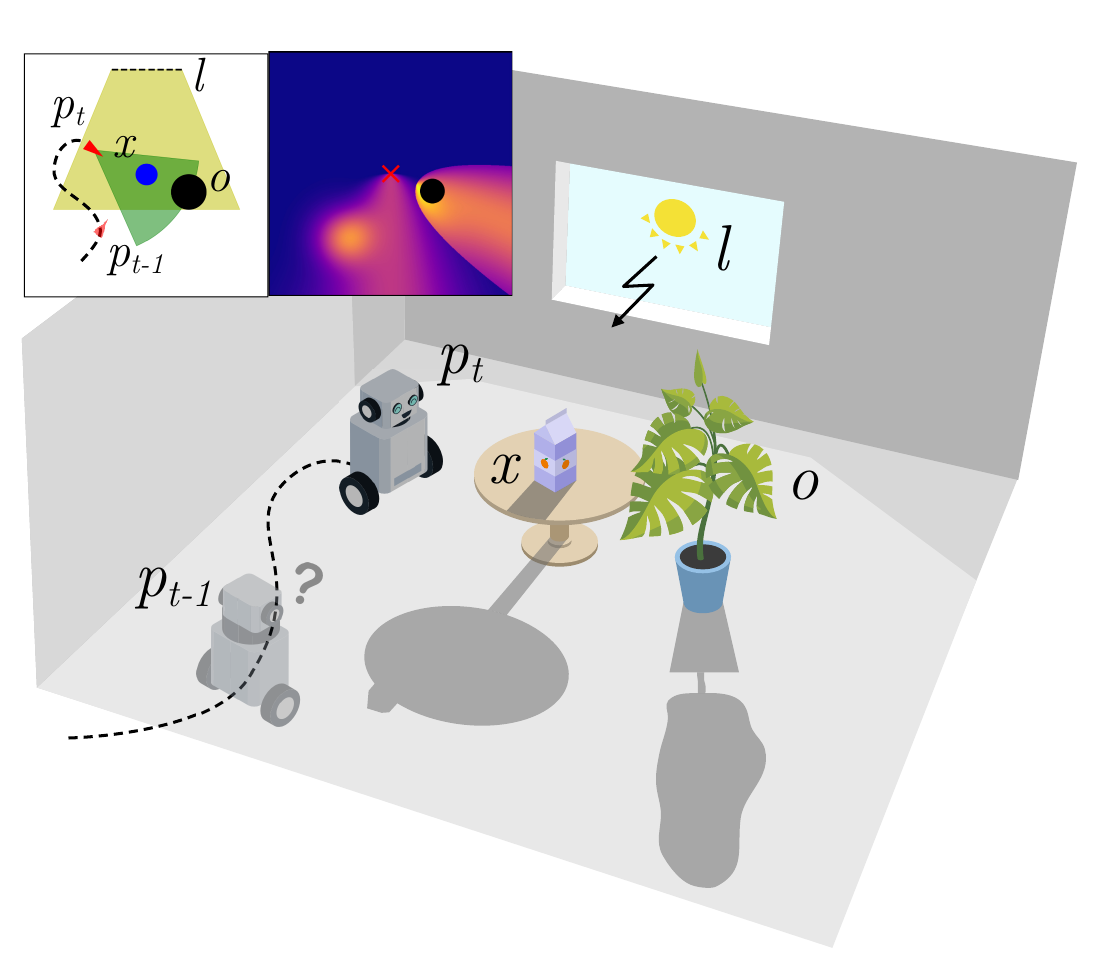}
    \caption{Overview of the active perception task where a robot moves to poses $\mathbf{p}$ to observe a target object $\mathbf{x}.$ A top-down view can be observed in the upper left corner. By integrating perceptual factors such as occluders $\mathbf{o},$ light sources $\mathbf{l},$ and previous poses $\mathbf{p}_{t-1},$ that condition potential observations shown in the heat map, our method enables accurately estimating the utility/value of potential viewpoints (blue is higher), which improves active perception performance}
    \label{fig:teaser}
    \vspace{-10pt}
\end{figure}

\section{Introduction} 
\label{sec:intro}

Scene understanding is a core prerequisite for autonomous robotic planning and decision-making. %
In various tasks, an important capability is active perception, whereby a robot actively moves a sensor in order to acquire relevant information from the environment~\cite{bajcsy1988active}.
Advances in computer vision and deep learning have substantially improved scene understanding from visual data~\cite{apatheodorou2023findobj,liu2022active,dharmadhikari2023semexpl,dang2018autonomous}.
An open challenge is deciding how to interpret and reason about such sensor information, which depends on the environment's appearance, to improve global understanding and active perception \cite{velez2012modelling, tchuiev2018inference}.

Traditional perception models~\cite{kantaros2021sampling, best2019dec} used for active perception assume that observations are independent given the robot pose and a target state to estimate, e.g., the position or the class of an object. 
This assumption falls short for complex sensors such as vision-based neural networks~\cite{xu2020fast}. 
These sensors extract information about visually-based high-level features of the scene. Consequently, they have a strong dependency on environmental characteristics, such as the appearance of objects, partial occlusions or back lighting, and show important spatial correlations due to appearance similarities. 
Disregarding such correlations spoils the estimation about the information gain of candidate observations, hindering the performance of active perception solutions.

To overcome this limitation, our main contribution is a new perception model that improves the estimation of the information gain provided by candidate viewpoints. 
This is achieved by modeling an arbitrary number of effects with general functions termed \textit{perceptual factors}.
These functions are combined to obtain a perceptual cost for the viewpoint that effectively weights the estimated information gain by increasing the measurement uncertainty when the information is redundant or affected by adverse environmental effects. The estimation is then used to evaluate the potential new viewpoints in active perception planning in order to acquire the most informative measurements. To showcase the generality of our approach, we also propose examples of perceptual factors relevant to robotic perception tasks.

 To validate the approach, we apply our active perception framework in a simulated environment that reproduces environmental effects for the tasks of object pose estimation and semantic classification. Finally, in order to demonstrate the practicality of our approach we deploy the active perception framework in a real-world scenario using a ground robot to perform active object classification in the presence of occluders and a light source. The results show how accounting for perceptual factors enhances state estimation quality. This leads to improved consistency in uncertainty and error management, ultimately boosting the robustness and efficiency of information gathering in simulated and real scenarios.

\section{Related Work}
\label{sec:related}

\subsection{Active Perception}

Considering how to actively select measurements that provide relevant information has been a long-standing problem in robotics. With a focus on the environment, exploration, and active reconstruction methods often use volumetric gain measurements that favor the observation of new geometry \cite{delmerico2018comparison, song2021view, schmid2020efficient}. Other methods quantify the information gain in a more advanced fashion by including information about the angle between the observations, or surface normals \cite{hepp2018plan3d, david2022sweep}. While these methods reason about how measurements are affected by the environment's appearance, they are exclusively based on geometry.

Traditional active perception approaches often assume measurement independence given the robot pose and the state to estimation to predict potential measurement values~\cite{kantaros2021sampling, popovic2020informative, best2019dec}. Consequently, the expected information gain of a viewpoint is purely a function of its location and the estimated state~\cite{kantaros2021sampling, best2019dec}. These are still the base for many active perception methods that reason on high-level semantic information \cite{liu2022active, asgharivaskasi2023semantic}. Extending from traditional methods, new perception models have been proposed. For instance, works based on unmanned aerial vehicles (UAVs)~\cite{popovic2020informative,qingqing2020towards,meera2019obstacle} typically use empirically determined altitude-dependant sensor model. While these approaches can be practical, by not considering correlations with the environment, they neglect the reduced utility that redundant measurements and measurements affected by other factors have.

Velez et al. \cite{velez2012modelling} first considered how measurements are correlated through the environment and approximated its effects by correlating measurements with previous viewpoints for object detection using Gaussian Processes. Several works stem from this formulation to incorporate planning \cite{teacy2015observation,tchuiev2023epistemic} and new models that include additional information about the viewpoint dependency in neural networks \cite{tchuiev2018inference, feldman2018bayesian}. In the present work, we consider a more general approach where we do not assume that all the influence in the environment is correlated with previous poses. Instead, we allow for more general models that include the influence of variables in the environment external to the object itself such as occluding objects, and light conditions.

\subsection{Environmental modeling}

We also consider works that model the environment despite they are not focused on active perception. Some methods apply learning methods to overcome the effects of the environment such as difficult light conditions \cite{kuznetsova2020using,xu2020fast}, or to remove occluding objects \cite{cheng2018reinforcement}. Rather than hardly reasoning about inadequate measurements to extract information, we focus on exploiting the positioning ability of robotic platforms to capture better and more informative measurements. Some of these environmental effects are solved by the application of specific solutions, such as in the case of occlusions, which are commonly considered in navigation \cite{schlegel2021blindspot, gilhuly2022occlusions} and manipulation \cite{morrison2019picking, menon2022viewpoint}. While we aim to model diverse and complex relations in perception models, we follow a general, synergetic approach that unifies them under the same framework and allows us to integrate them into planning to obtain the most informative viewpoints.
\section{Problem Formulation}

Consider a robot moving in an environment equipped with a visual sensor to gather data. We denote $\mathbf{p}_t$ as the robot viewpoint at time $t$.
The objective of the robot is to estimate the state of one or more targets, denoted by $\mathbf{x}_t$. This information can be of any kind, e.g., metric (target location) or semantic (target class).
To simplify the notation, we assume the target's state does not change over time and drop the dependence on $t$, but this is not an actual limitation of our problem setup.

\begin{figure}[t]
    \centering
    \begin{subfigure}[b]{0.175\textwidth}
         \centering
         \includegraphics[width=\textwidth]{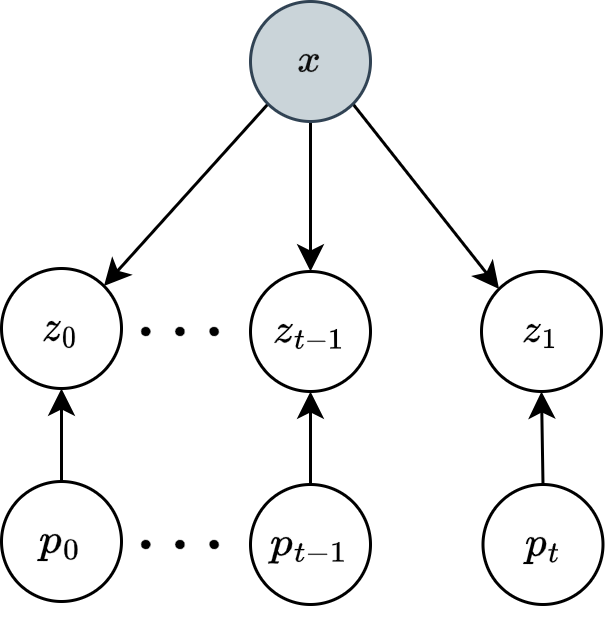}
         \caption{Traditional}
         \label{fig:traditional}
     \end{subfigure}
      \begin{subfigure}[b]{0.22\textwidth}
         \centering
         \includegraphics[width=\textwidth]{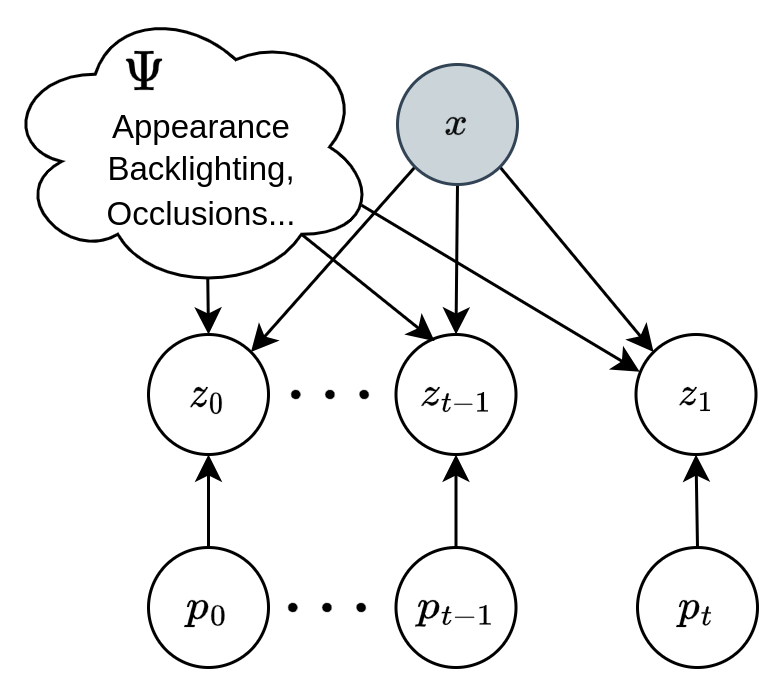}
         \caption{General perception model}
         \label{fig:ours}
     \end{subfigure}
    \caption{Perception models considering different measurement dependencies. Traditional models (a) assume measurements are independent of each other. In reality (b), they are influenced by diverse environmental variables $\Psi$.}
    \label{fig:perceptionmodels}
\end{figure}

The robot has a perception algorithm available, such as a vision-based neural network, used to obtain measurements, denoted $\mathbf{z}_t$, of the target $\mathbf{x}$ at the viewpoints.
Differently from classical setups, where the measurements are assumed to be conditionally independent (\reffig{fig:traditional}), we consider that the value of $\mathbf{z}_t$ also depends on other environmental factors, $\Psi$ (\reffig{fig:ours}),
\begin{equation}
    \mathbf{z}_t = g(\mathbf{x}, \mathbf{p}_t,\Psi).
\end{equation}
The measurements are used to infer a probability distribution over the target state, $\hat{\mathbf{x}}_t\equiv p(\mathbf{x}\,|\, \mathbf{z}_{0:t}, \mathbf{p}_{0:t},\Psi).$

In this setting, we consider an active perception task. Our goal is to move the robot sensor to maximize the information gathered from the environment. To this end, we leverage an informative path planning (IPP) algorithm, which involves finding the policy $\pi = (\mathbf{p}_1, \dots, \mathbf{p}_N), N>0$ that optimizes a desired quality criterion, $I$, over the state estimation,
\begin{equation}
\label{Eq:IPP_problem}
\begin{split}
        &\pi^\star = \argmax_{\pi \in \Pi} I\bigl(\hat{\mathbf{x}}_N\bigr), \\
        \text{ s.t. } 
        &\mathbf{z}_t = g(\mathbf{x}, \mathbf{p}_t,\Psi),\\
        &\hat{\mathbf{x}}_t = h(\hat{\mathbf{x}}_{t-1}, \mathbf{p}_t, \mathbf{z}_t),\\
        &C(\pi) \leq B,
\end{split}
\end{equation}
where $h(\cdot)$ is the estimation algorithm used for perception, $C(\pi)$ is the policy cost, $B \geq 0$ is the mission budget, e.g., traveled distance, and $\Pi$ is the set of admissible viewpoints or paths of length $N$.

Our key contribution is a model for $h(\cdot)$ in Eq.~\eqref{Eq:IPP_problem} which captures $\Psi$ in a general, simple and efficient way. Our new model enables accurately estimating the information gain associated with potential viewpoint candidates, thereby improving active perception performance.

\section{Approach} %
\label{sec:main}

\subsection{Estimation Model}
We restrict our analysis of $h$ to Bayesian recursive estimation algorithms. 
Considering the effect of $\Psi$ in the recursive estimation problem results in the following:
\begin{equation} %
\label{eq:env_model}
    h(\hat{\mathbf{x}}_{t-1}, \mathbf{p}_t, \mathbf{z}_t) = 
    \frac{p(\mathbf{z}_t | \mathbf{x}, \mathbf{p}_t, \Psi) 
    \hat{\mathbf{x}}_{t-1}}{p(\mathbf{z}_t | \mathbf{p}_t, \Psi)}.  
\end{equation}

Accounting for all perceptual properties of the environment and their influence on the measurement likelihood in Eq.~\eqref{eq:env_model} is a very complex problem.
To address this, we propose a map representation where multiple factors can be efficiently combined to approximate these effects.
We then discuss how to use this map in different fusion mechanisms to compute the posterior estimation and provide examples of different factors.

\subsection{Perceptual Maps}
To model the influence of $\Psi$ in the potential viewpoints we introduce a set of functions, $f(\mathbf{p}_t, \theta) \in [1, \infty),$ that we call \textit{perceptual factors}, where $\theta$ are parameters that can be used to describe different influences.
All the perceptual factors can be combined into a joint \textit{perceptual map} that approximates the total influence simply by multiplying them, 
\begin{equation}
    \hat{\Psi}(\mathbf{p}_t, \theta) := \prod_j f_j(\mathbf{p}_t, \theta_j),
\end{equation}
where $\theta_j\subseteq \theta$ are the parameters describing each perceptual factor $j$. Therefore, for each robot pose $\mathbf{p}_t,$ the perceptual map function contains the \textit{perceptual cost} associated with that viewpoint. Similar to other kinds of maps, the perceptual map can be built with prior information about the environment or updated online during the mission by including newly discovered information.

In the following, we explain how the perceptual cost can be applied to continuous variables using a standard Extended Kalman Filter (EKF) formulation and a categorical fusion for discrete classification. These examples represent typical robotic perception problems using image processing, i.e., object pose estimation and semantic classification. 

\subsubsection{Extended Kalman Filter (EKF)} 
Assuming Gaussian distributions for all the probabilities in Eq.~\eqref{eq:env_model}, and recalling that we assume a static state (i.e., no need for prediction), the EKF computes the mean $\mathbf{\hat x}_{t}$ and covariance $\mathbf{P}_t$ of the posterior distribution by
\begin{equation}
\label{Eq:EKF}
    \begin{split}
        \mathbf{K}_t &= \mathbf{P}_{t-1} \mathbf{H}_t^T (\mathbf{R}_t + \mathbf{P}_{t-1} \mathbf{H}_t^T)^{-1}, \\
        \mathbf{\hat x}_{t} &= \mathbf{\hat x}_{t-1} + \mathbf{K}_t (\mathbf{z}_t - \mathbf{H}_t \mathbf{\hat x}_{t-1}), \\
        \mathbf{P}_{t} &= (\mathbf{I} - \mathbf{K}_t \mathbf{H}_t) \mathbf{P}_{t-1},        
    \end{split}
\end{equation}
where $\mathbf{H}_t$ is the Jacobian of the observation function linearized around the state estimate, and $\mathbf{R}_t$ is the sensor covariance matrix. 

Our proposed framework includes the perceptual cost as part of the sensor covariance matrix via multiplication,
\begin{equation}
\label{Eq:R-perceptual}
    \mathbf{R}_t^\Psi = \mathbf{R}_t  \hat{\Psi}(\mathbf{p}_t\,, \theta),
\end{equation}
where the original value of $\mathbf{R}_t$ is obtained from the sensor calibration in standard conditions.
Since $f(\mathbf{p}_t, \theta)\ge 1,$ this modification preserves the principles of the EKF, such that $\mathbf{R}_t^\Psi$ is a symmetric and positive semi definite matrix, but also encodes potential environment uncertainties.
A large value of the perceptual map at that viewpoint increases $\mathbf{R}_t^\Psi,$ weighing down the effect that the measurement would normally have.
Note that our perceptual factors can thus be applied to any existing IPP algorithm based on the EKF \cite{kantaros2021sampling, popovic2020informative}.

\subsubsection{Categorical Estimator}
Perceptual factors are especially beneficial in deep-learning based semantic vision algorithms since they strongly depend on the measurement fusion process.
In the case of semantic classification, $\hat{\mathbf{x}}_t$ and $\mathbf{z}_t$ are described by two probability vectors, where $\hat{x}_{i,t}$ is the probability of $\mathbf{x}$ being of class $i$ (respectively $z_i$).
Similarly to our previous work~\cite{morillacabello2023}, we model the categorical semantic fusion as a Dirichlet distribution, where the concentration parameter expresses the relative importance between the prior and likelihood distributions to compute the posterior:
\begin{equation} \label{Eq:SemanticFusion}
\hat{x}_{i,t} \propto
    \left(\hat{x}_{i,t-1}\right)^{\frac{\bar{\alpha}_{i,t-1}}{\bar{\alpha}_{i,t}}} \left( z_i \right)^{\frac{\alpha_{i,t}}{\bar{\alpha}_{i,t}}},
\end{equation}
where $\bar{\alpha}_{i,t-1}$ and $\alpha_{i,t}$ are the concentration parameters of the prior and the measurement respectively, and $\bar{\alpha}_{i,t}=\max(\bar{\alpha}_{i,t-1},\alpha_{i,t})$ is a normalizing term.
We define
\begin{equation}
\label{Eq:FusionWeighted}
  \alpha_{i,t} = \dfrac{1}{\hat \Psi(\mathbf{p}_t\,, \theta)} \in (0,1], \hbox{ and }
  \bar{\alpha}_{i,t} = 1, \forall t,
\end{equation}
to similarly reduce the influence of measurements with an associated large perceptual cost.
Similar to the EKF, the way in which we introduce the perceptual factors in~\eqref{Eq:FusionWeighted} permits using existing planners to solve Eq.~\eqref{Eq:IPP_problem} \cite{asgharivaskasi2023semantic, serra2023activeclassification}.

\begin{figure}[t]
    \centering
    \includegraphics[width=0.95\columnwidth]{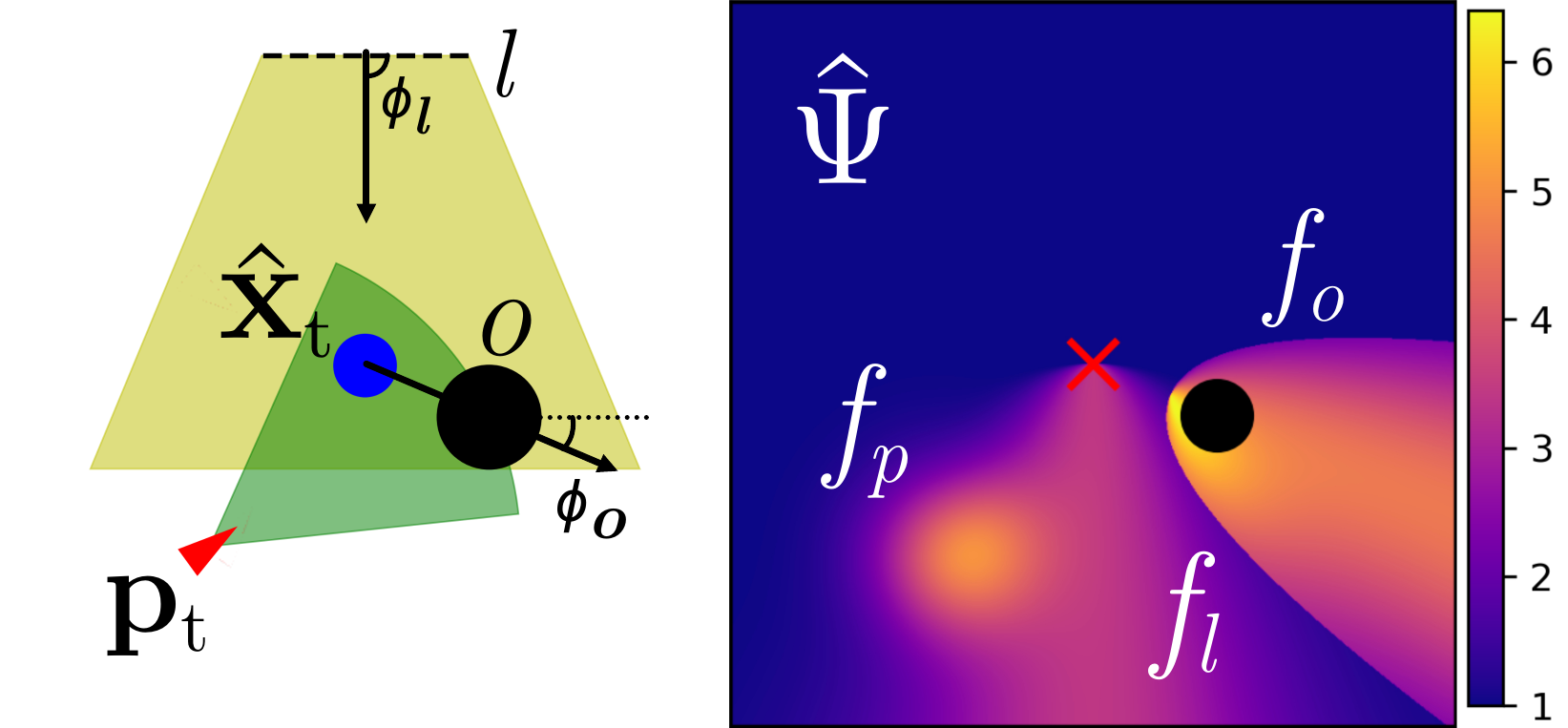}
    \caption{Perceptual map combining three different perceptual factors. (Left) Environment with one target (blue circle and estimation $\hat{\mathbf{x}}_t$), one obstacle ($\mathbf{o}$), one light source ($l$) and the last viewpoint ($\mathbf{p}_{t}$).
    (Right) Perceptual map of the target combining the three factors: occlusions ($f_o$), back lighting ($f_l$) and past viewpoints ($f_p$). Larger values in the map identify less informative viewpoints.}
    \label{fig:factors}
\end{figure}

\subsection{Examples of Perceptual Factors}\label{subsec:perceptual_factors}
Our formulation offers a general framework that can support different mathematical representations of perceptual factors. Examples include analytically differentiable functions efficient for gradient-based planning or learned functions capturing intricate interactions with the environment.
We introduce three hand-crafted examples relevant to robotic applications that model direct effects of the scene and redundancy due to local appearance. 

Partial occlusions and back lighting are two environment elements that have a strong influence on the outcome of deep-learning vision algorithms.
To keep computation simple and efficient, we model both factors with bounded quadratic, functions translated and rotated, with parameters $\theta=\{\delta, w\}$ defining the strength of the influence and the width of the parabola respectively.

For a 2D environment, given an obstacle located at $\mathbf{o}=(o_x,o_y)$, the resulting factor is
\begin{equation}
\label{Eq:Obstacle_Factor}
\begin{split}
    f_{o}(\mathbf{p}_t\,, \mathbf{o}) &= \begin{cases} 1 + \delta \exp{\frac{-x'^2}{y'}} &\text{ if } \frac{x'^2}{\sigma}  < y', \\ 1 &\text{ otherwise } \end{cases} \\
    x' &= (p_x-o_x) \cos{(\phi)} - (p_y-o_y) \sin{(\phi)} , \\
    y' &= (p_x-o_x) \sin{(\phi)} + (p_y-o_y) \cos{(\phi)} ,
\end{split}
\end{equation}
where $(p_x, p_y)$ is the position in $\mathbf{p}_t$ and $\phi=\arctan(\hat{y}-o_y, \hat{x}-o_x)$ with $(\hat{x},\hat{y})$ the target position.
The width is equal to the sum of the obstacle and target sizes in order to free the vision and avoid partial occlusions. The 3D case can be computed in a similar fashion.

The back lighting generated by an directional light source is very similar. To compute $f_{l}(\mathbf{p}_t\,, \theta)$ we follow the same computations as in Eq.~\eqref{Eq:Obstacle_Factor}, changing the origin of the parabola to the target, $\hat{\mathbf{x}}_t,$ and defining $\phi$ as the direction of the light.
The width, in this case, will depend on the light diffusion, the more concentrated the light source is in its direction, the narrower its effect. 

The fact that measurements are captured by cameras from the scene makes the information of nearby viewpoints redundant due to similar appearance and does not contribute to improving the state estimate. 
We introduce a second type of perceptual factor that aims to reduce the influence of subsequent measurements captured at similar viewpoints:
\begin{equation}
    f_{r} = 1 + \delta \exp\left(\frac{-\|\mathbf{p}_t-\mathbf{p}_{i}\|_2^2}{2\sigma^2}\right ).
\end{equation}

As an illustrative example, \reffig{fig:factors} combines the presented factors into a perceptual cost map, which corresponds to the physical environment shown in \reffig{fig:teaser}.

\section{Experimental Results}
\label{sec:exp}

\begin{figure}[t]
    \centering
    \begin{subfigure}[b]{0.215\textwidth}
         \centering
         \includegraphics[width=\textwidth]{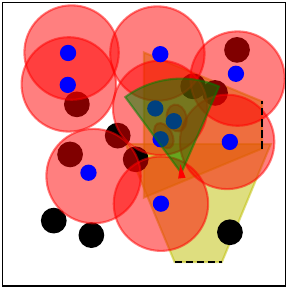}
       
         \label{fig:metric_sim}
     \end{subfigure}
   \begin{subfigure}[b]{0.215\textwidth}
         \centering
        \includegraphics[width=\textwidth]{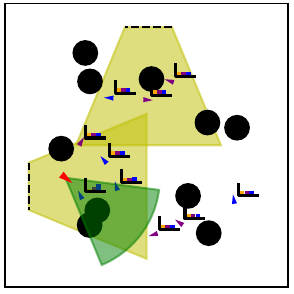}
         
         \label{fig:semantic_sim}
     \end{subfigure}
      
    \caption{Example of environments for the \textit{metric} (left) and \textit{semantic} (right) simulated experiments. A robot (red) moves to observe targets (blue). Occluders (black) and lights (yellow) may add noise to the measurements. The estimation covariance is shown in red for the \textit{metric} experiment. The classification histogram is shown for the \textit{semantic} experiment. The supplementary video shows some executions.}
    \label{fig:simulator_examples}
\end{figure}

\subsection{Simulated Experiments}
\label{SS:simulated_exp}

\begin{figure}[t]
    \centering
     \begin{subfigure}[b]{0.45\columnwidth}
         \centering
         \includegraphics[width=\textwidth]{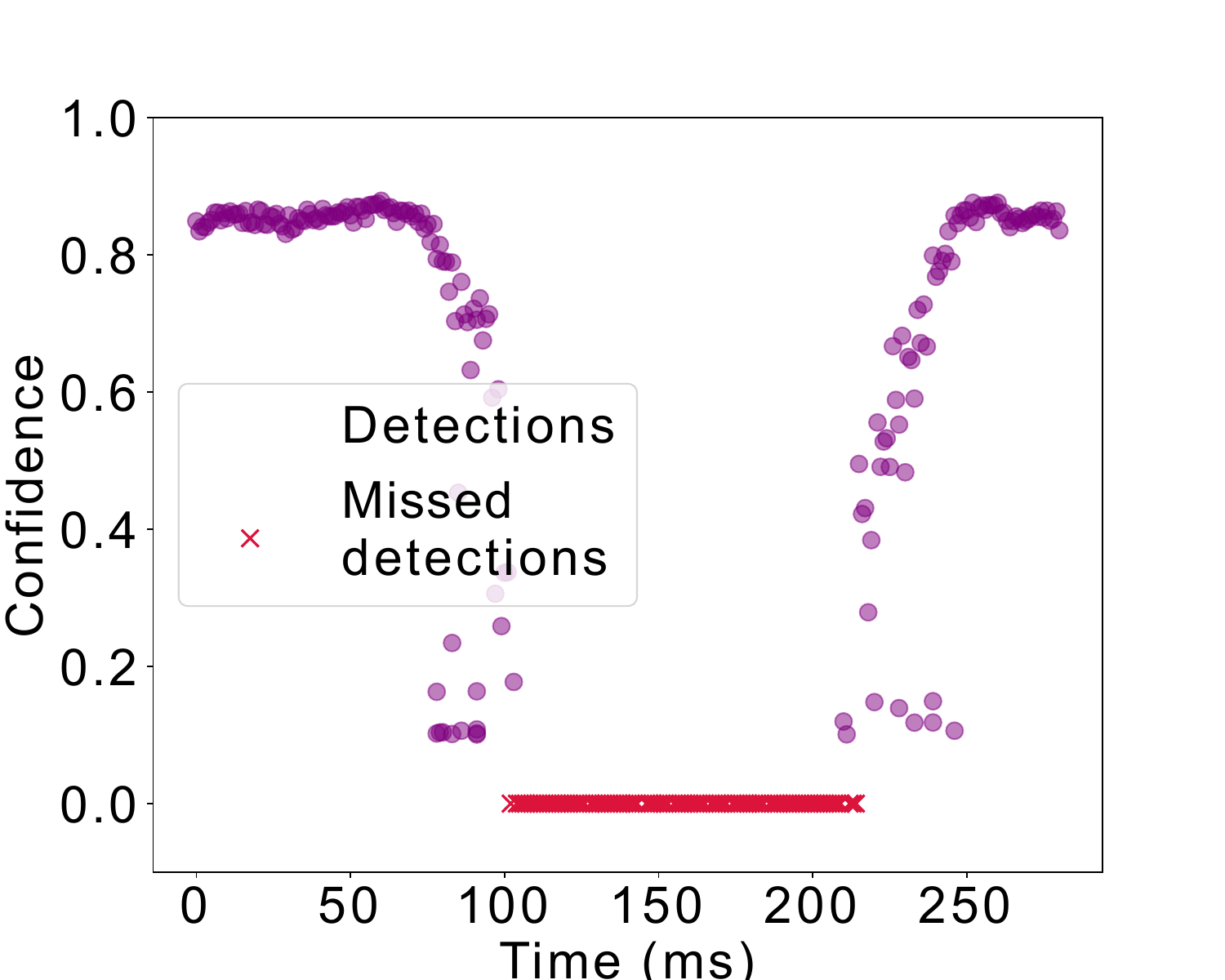}
         \caption{Occlusion effect}
         \label{fig:occlusion_model}
     \end{subfigure}
     \begin{subfigure}[b]{0.45\columnwidth}
         \centering
         \includegraphics[width=\textwidth]{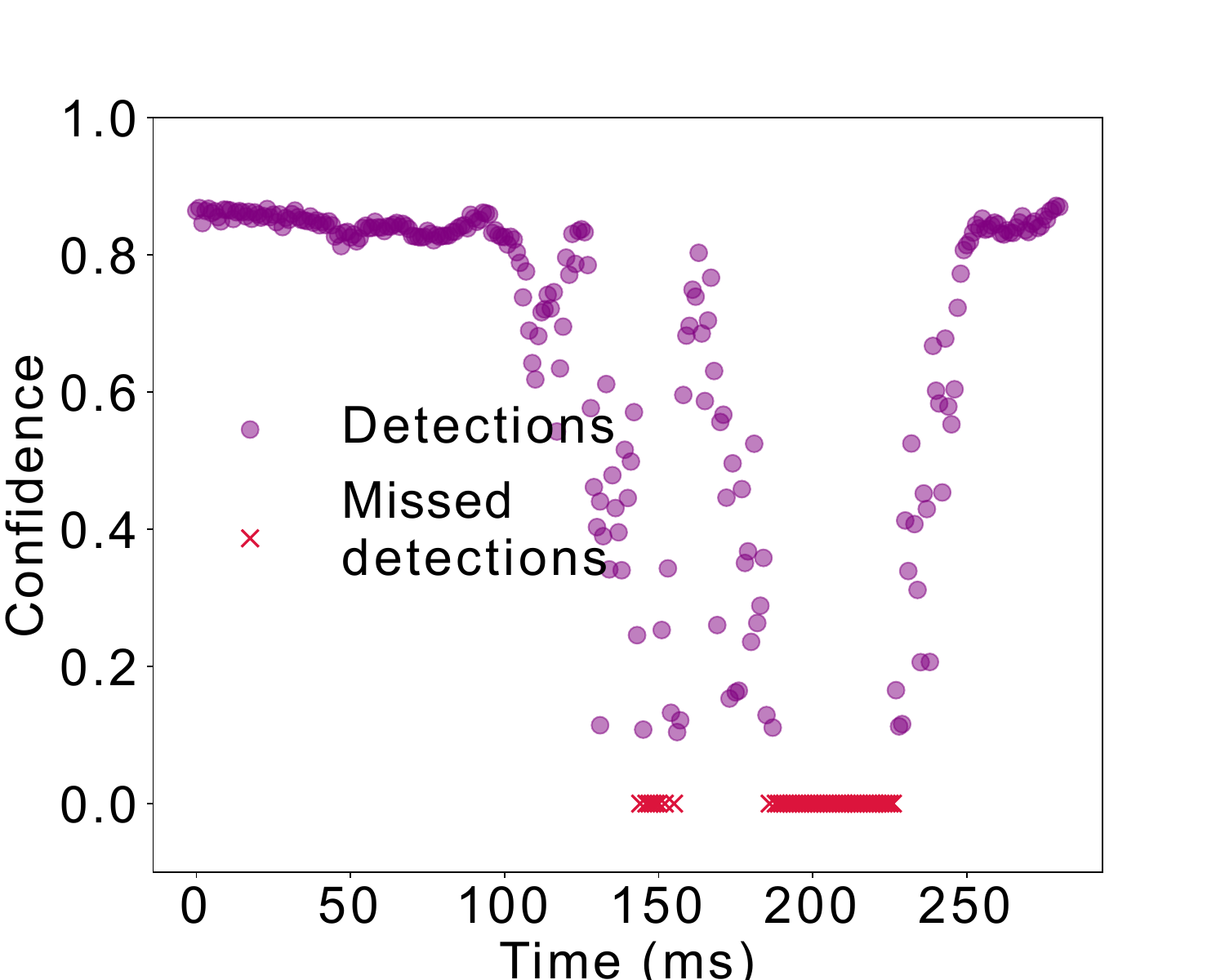}
         \caption{Light effect}
         \label{fig:light_model}
     \end{subfigure}
    \begin{subfigure}[b]{0.45\columnwidth}
         \centering
         \includegraphics[width=\textwidth]{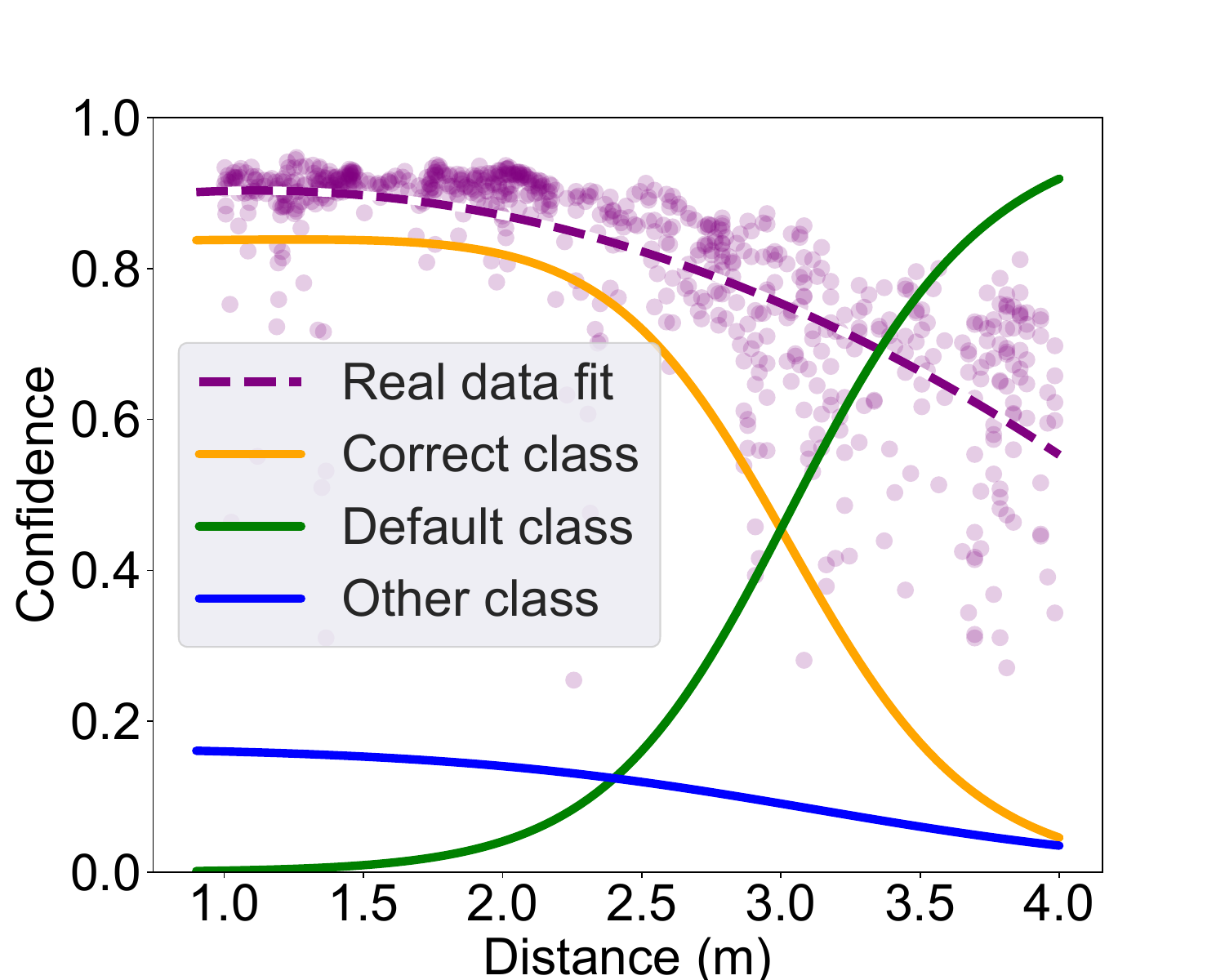}
         \caption{Real detector \\ data}
         \label{fig:distance_model}
     \end{subfigure}
       \begin{subfigure}[b]{0.45\columnwidth}
         \centering
         \includegraphics[width=\textwidth]{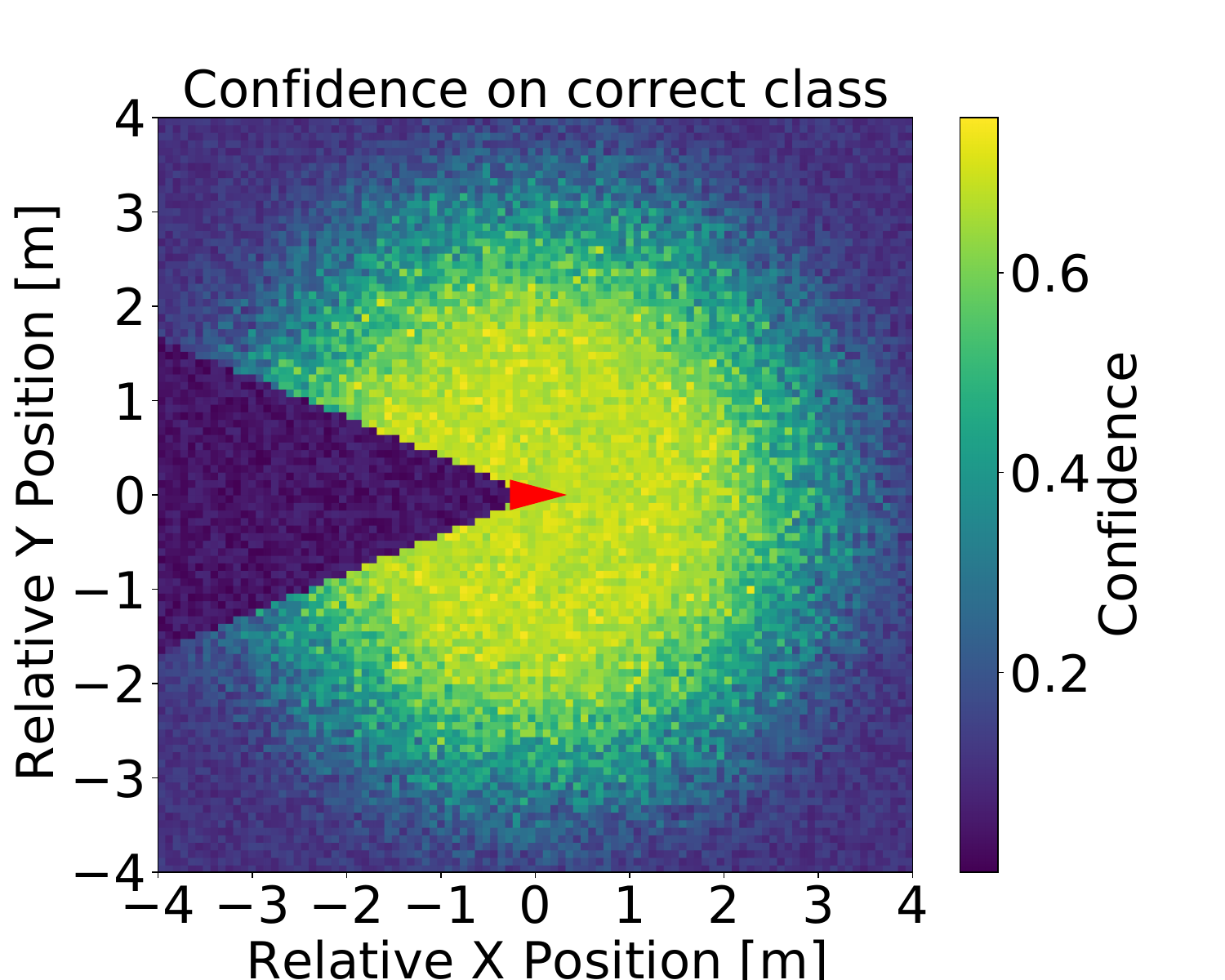}
         \caption{Simulated \\ semantic model}
         \label{fig:model_semantic}
     \end{subfigure}
    \caption{Characterization of real-world perception models used in our simulator. We capture real detections from a classifier (in purple) in order to model the effects of (a) occlusions and (b) back lighting. The capturing process is included in the supplementary video. We use the response to semantic detections to model (c) simulated confidence. Together with noise, and a region where the target is misclassified, we build (d) a simulated semantic model.}
    \label{fig:semanticsimulator}
    \vspace{-10pt}
\end{figure}

\textbf{Setup.} We evaluate our method in two active perception tasks: pose estimation, termed \textit{metric}, and semantic classification, termed \textit{semantic}.
We consider 50 randomly generated 2D $10m \times 10m$ environments, with $10$ targets to localize/classify, $10$ occluders, and $2$ light sources. \reffig{fig:simulator_examples} show examples of these environments. To simulate measurements that resemble the actual output of vision algorithms affected by environmental factors, we execute a characterization of the noise from a pre-trained object detector in different controlled experiments. \reffig{fig:occlusion_model} shows measurement confidences and missed detections when the target is incrementally occluded by a moving object. Similarly, \reffig{fig:light_model} exemplifies a light source generating back lighting when moving behind the object. Based on the captured data, we model our simulator to add noise to the measurements depending on which, and how many factors affect the measurement. Each partial occlusion adds noise incrementally with the occlusion amount. Finally, lights add a fixed increment when the target is between the light and the robot.

For the \textit{metric} problem, the robot is equipped with a range-bearing sensor. The sensor returns measurements corrupted with Gaussian noise with standard deviation for distance $\sigma_d = 0.3$ and bearing $\sigma_b = 0.1$. The standard deviation is scaled upon the effect of perceptual factors by $\gamma = [1,6]$. The additional noise also increases the probability of a missed detection.

For the \textit{semantic} experiment, the sensor returns classification confidences for each measured target. In this case, we also characterize the confidence response from the real object detector.
According to \reffig{fig:distance_model}, we model the distance effect by decaying confidence in the detected class and increasing noise, which results in outliers. We also add a region where the object is classified as the wrong class, resulting in the model shown in \reffig{fig:model_semantic}. Similarly to the \textit{metric} case, we include the perceptual factors' noise increment in the confidence noise and the probability of outliers.

\textbf{Baselines.} 
To carry out the perception tasks, we use Eq.~\eqref{Eq:IPP_problem} to decide which next viewpoints for the robot to take measurements. 
We consider the entropy of $\hat{\mathbf{x}}_t$ as the quality criterion, $I(\hat{\mathbf{x}}_t),$ the Euclidean distance for $C(\pi)$, the EKF in Eq.~\eqref{Eq:EKF} as estimation model for the \textit{metric} task and the categorical estimator of Eq.~\eqref{Eq:SemanticFusion} for the \textit{semantic} task.
We consider a planning horizon of $N=1$, use random sampling to generate $100$ candidate viewpoints with $B=2m$ and $2\pi$ rad for the orientation. Then, we choose the best candidate viewpoint in terms of $I(\hat{\mathbf{x}}_t)$ and move the robot to acquire a new measurement there. We repeat this process $50$ times in each simulation.

The IPP algorithm is designed in a simple way to emphasize the analysis of the perceptual factors, the core contribution of our paper.
We compare different versions of $\hat{\Psi}(\mathbf{p}_t,\theta)$ for the perceptual factors. The \emph{Basic} algorithm is the baseline in which perceptual factors are neglected, i.e., $\hat{\Psi}=1$ everywhere.
We also analyze each of the three proposed factors separately (\emph{Occlusions}, \emph{Light} and \emph{Previous Poses}), as ablation versions of our \emph{Complete} method. The perceptual factors are accounted in Eq.~\eqref{Eq:R-perceptual} and Eq.~\eqref{Eq:FusionWeighted} for the \emph{metric} and \emph{semantic} experiments respectively. For modeling these factors, we use the proposed functions in \refsec{subsec:perceptual_factors}. The influence values and widths for occluders, lights, and correlation with previous poses are $\delta_o = 3$, $\delta_l = 2$, $w_l = 3$, $\delta_p = 3$, and $\sigma_p=0.1$ respectively.

\textbf{Evaluation metrics.} Since the analyzed methods have different measurement models, in terms of their uncertainties, the amount of information gain of each method, $I(\hat{\mathbf{x}}_t),$ is not a representative metric to compare them. 
Moreover, entropy reduction does not necessarily imply a good estimation in the presence of very noisy measurements.
Instead, our evaluation focuses on the consistency of the estimation, understood as how the uncertainty represents the true error of the estimator.

In the case of the EKF, we use the Normalized Estimated Error Squared (NEES) averaged over all targets, defined as
\begin{equation}
    NEES = (\hat{\mathbf{x}}_t - \mathbf{x})^T \mathbf{P}_t^{-1} (\hat{\mathbf{x}}_t - \mathbf{x}).
\end{equation}
NEES values closer to the target dimension ($2$ in our experiments) indicate better filter estimation in terms of consistency.
We also report Root Mean Squared Error (RMSE) of the position estimation with respect to ground truth.

In the case of the categorical semantic fusion, we evaluate the average confidence of the true class, which represents the confidence of the posterior and also encodes the concept of consistency over the prediction. Additionally, we report the overall accuracy of the classification as the percentage of correctly classified targets.

\begin{figure}[t]
    \centering
    \begin{tabular}{cc}
        \includegraphics[width=0.45\columnwidth]{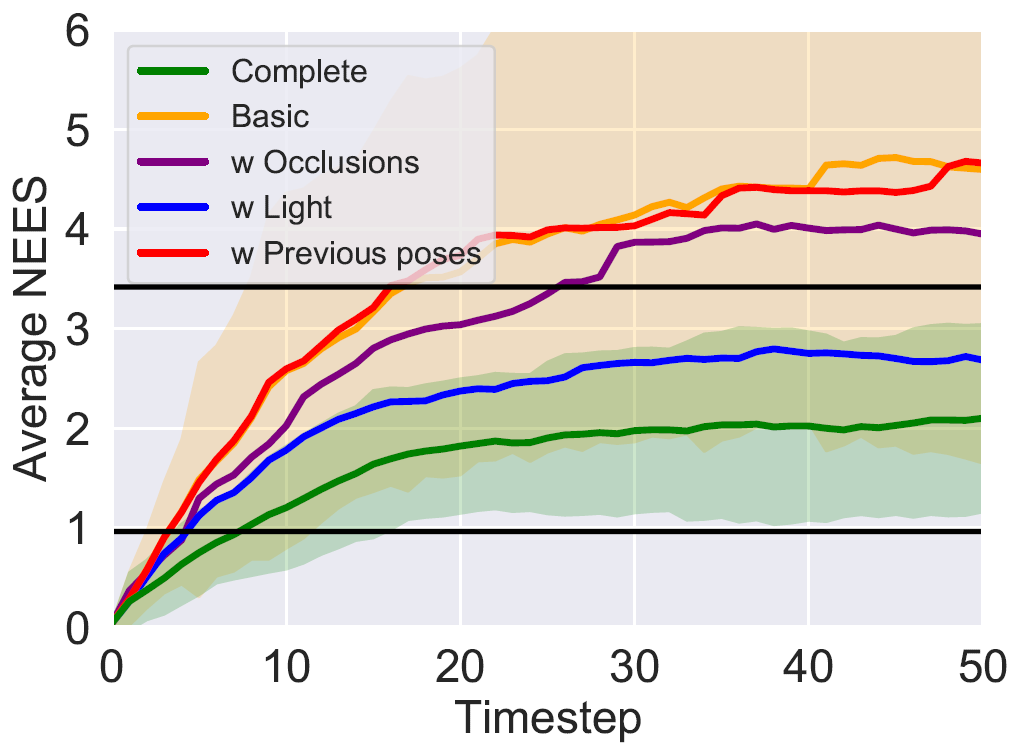} &
        \includegraphics[width=0.45\columnwidth]{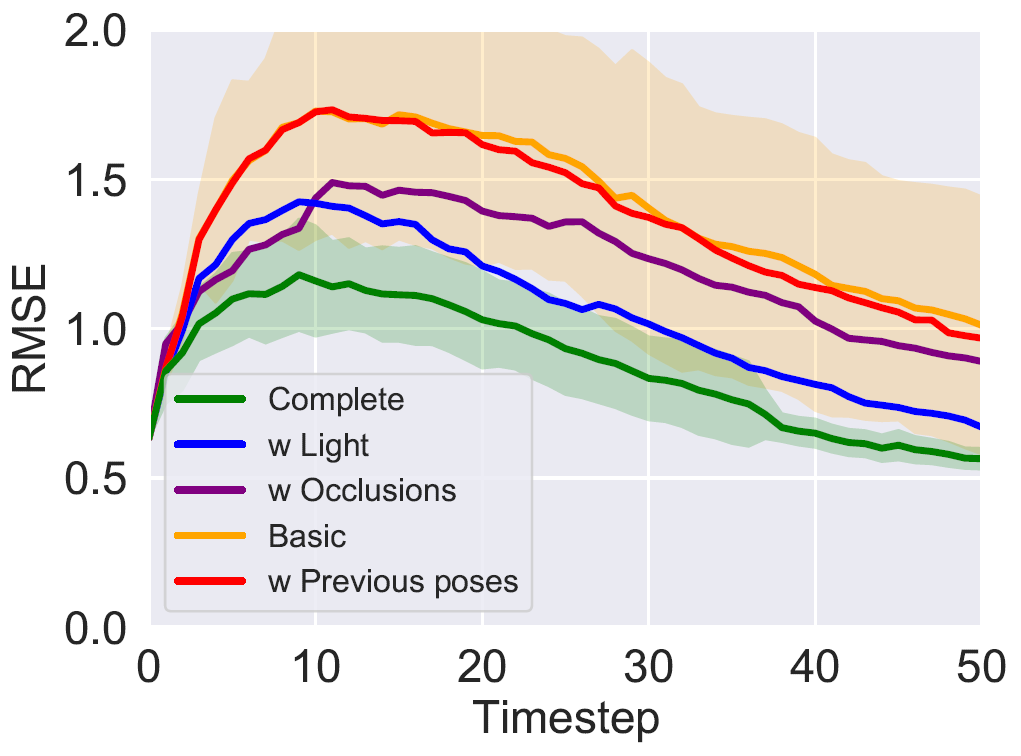} \\
        \includegraphics[width=0.45\columnwidth]{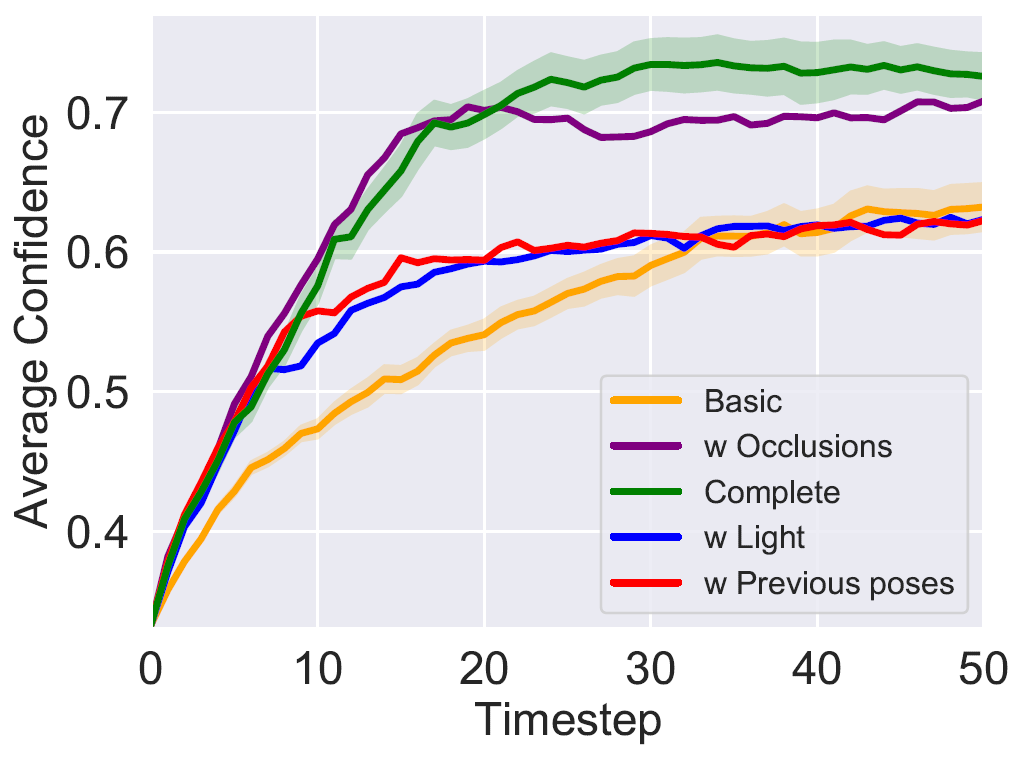} &
        \includegraphics[width=0.45\columnwidth]{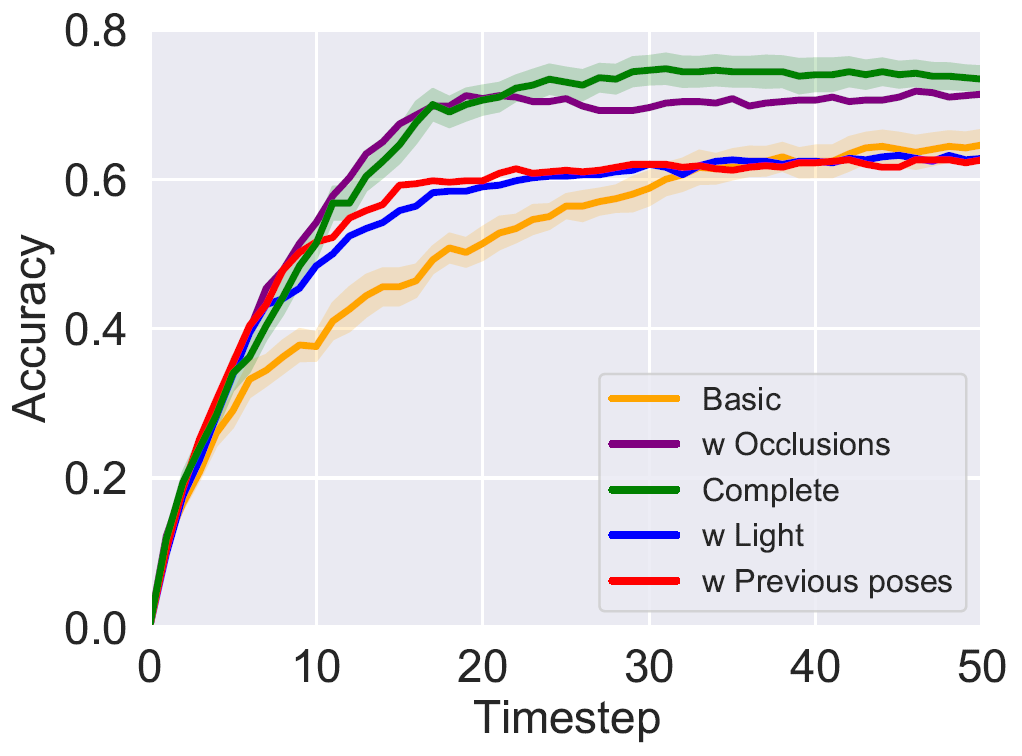}
    \end{tabular}
         \label{fig:semantic_acc}
    \caption{Averaged results (50 experiments) for the \textit{metric} (left) and \textit{semantic} (right) experiments. Introducing perceptual factors reduces average NEES and RMSE. Additionally, by accounting for the noise induced by perceptual factors, our method stays within the confidence interval for a significance level of $0.05$ (horizontal black lines), thus ensuring consistency. In the same way, we increase average confidence and accuracy by reducing the influence of outliers.}
    \label{fig:metric_graphs}
    \vspace{-10pt}
\end{figure}

\begin{figure}[t]
    \centering
    \begin{subfigure}[b]{0.235\textwidth}
         \centering
         \includegraphics[width=\textwidth]{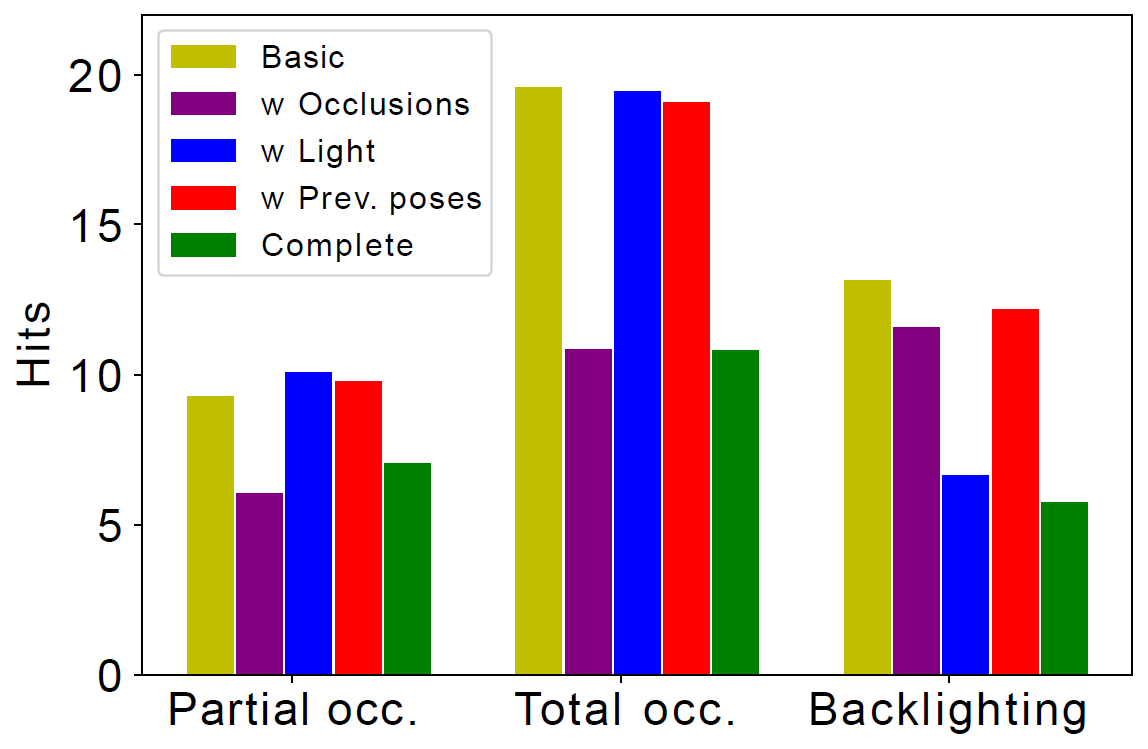}
       
         \label{fig:hist_metric}
     \end{subfigure}
   \begin{subfigure}[b]{0.235\textwidth}
         \centering
        \includegraphics[width=\textwidth]{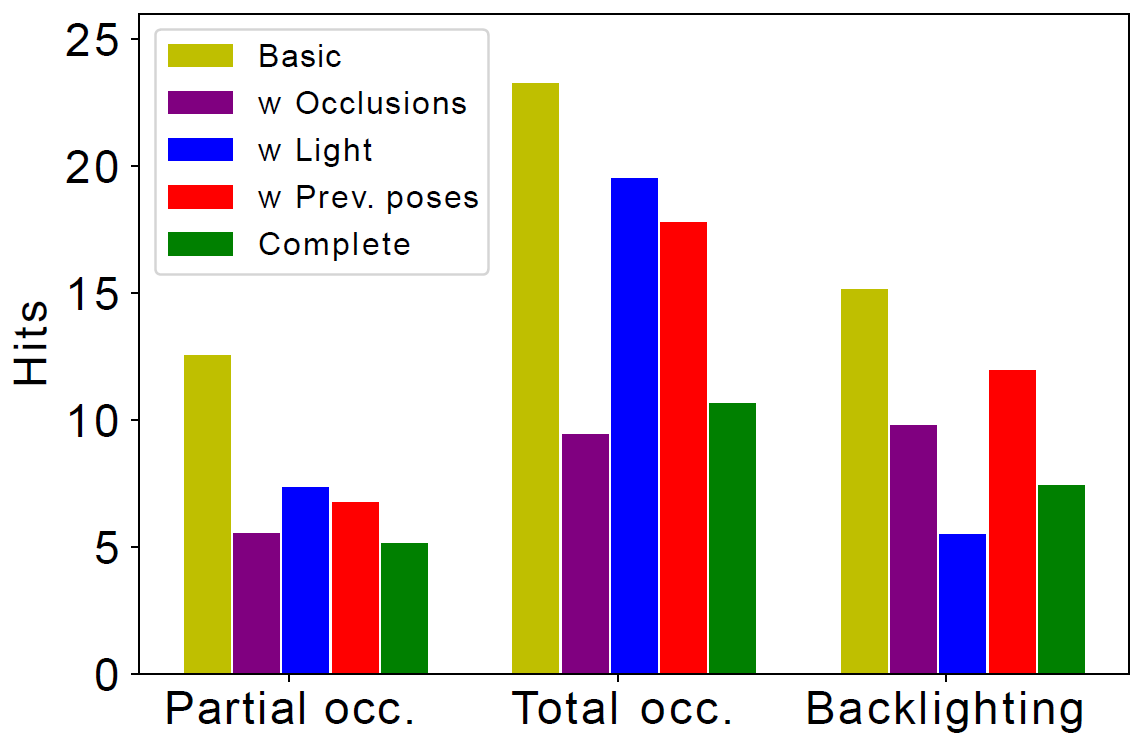}
         
         \label{fig:hist_semantic}
     \end{subfigure}
      
    \caption{Numbers of perceptual factors hit in the (left) \textit{metric} and (right) \textit{semantic} experiments. When we introduce perceptual factors, the planner estimates lower information value in the viewpoints they affect, guiding the robot to avoid these locations.}
    \label{fig:factors_hit}
\end{figure}

\textbf{Results.} As illustrated in \reffig{fig:metric_graphs}, the consistency of the estimation for the basic method increases during the mission execution beyond accepted levels of confidence. This is primarily attributed to noisy measurements resulting from partial occlusions and lighting conditions. Our method addresses these challenges via two significant enhancements. First, it selects measurement points that avoid these effects, as shown in \reffig{fig:factors_hit}. Second, our method reduces the confidence of noisy measurements during the fusion process. This diminishes their influence and reduces the estimated uncertainty, thereby ensuring consistent stability around the desired NEES value, which is $2$. These effects are also reflected in the RMSE graph, where our method effectively minimizes estimation errors by mitigating the impact of outliers on state estimation. 

For the \emph{semantic} problem, by including all the perceptual factors we obtain more confident measurements and a better accuracy in the classification (\reffig{fig:metric_graphs}). Similarly, the number of bad measurements, affected by occlusions or light is lower than the baseline (\reffig{fig:factors_hit}).

\begin{figure}[t]
    \centering
    \begin{subfigure}[b]{0.33\columnwidth}
         \centering
         \includegraphics[width=\textwidth]{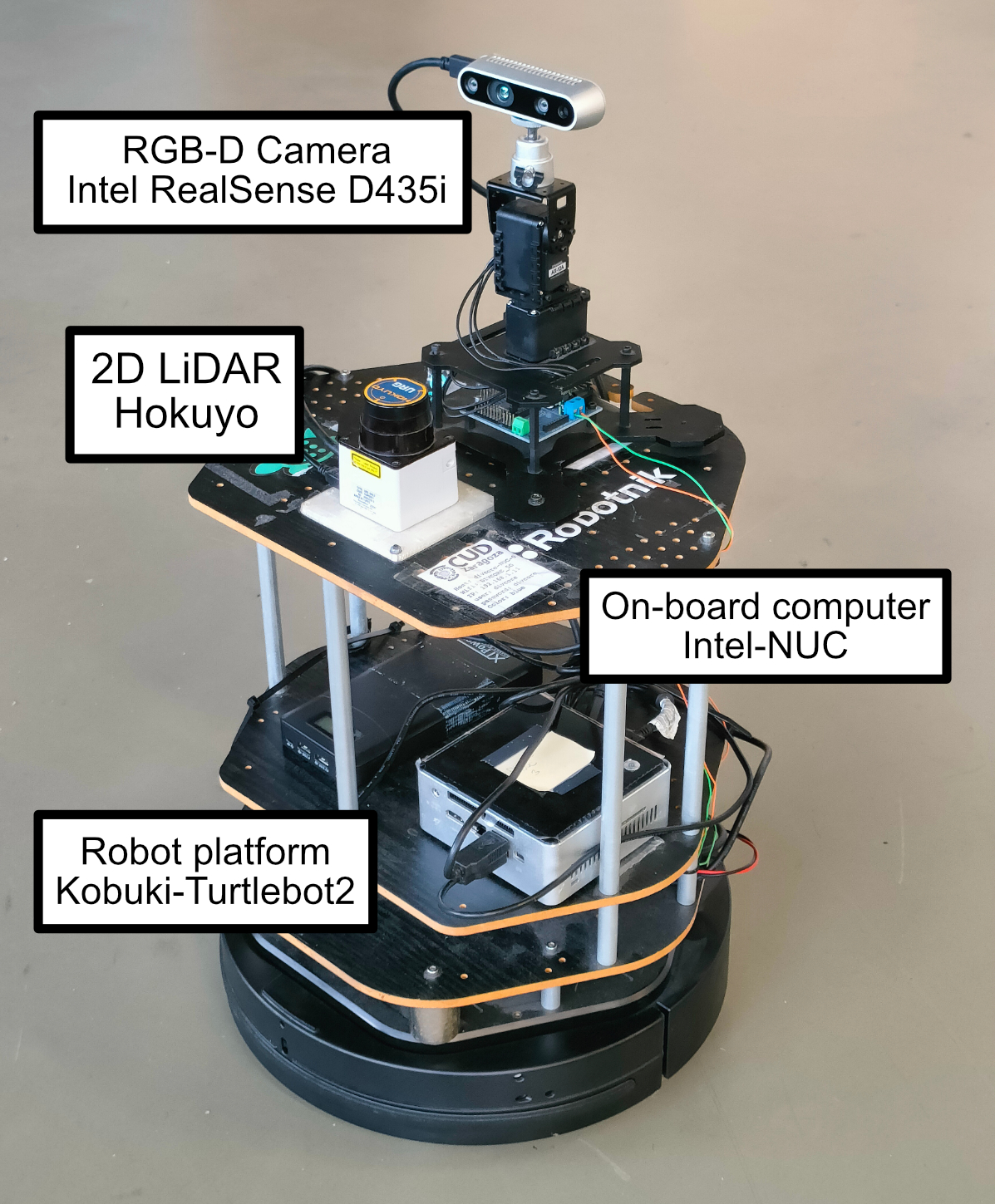}%
     \end{subfigure}
    \begin{subfigure}[b]{0.285\textwidth}
         \centering
         \includegraphics[width=\textwidth, trim={0 0.335cm 0 0},clip]{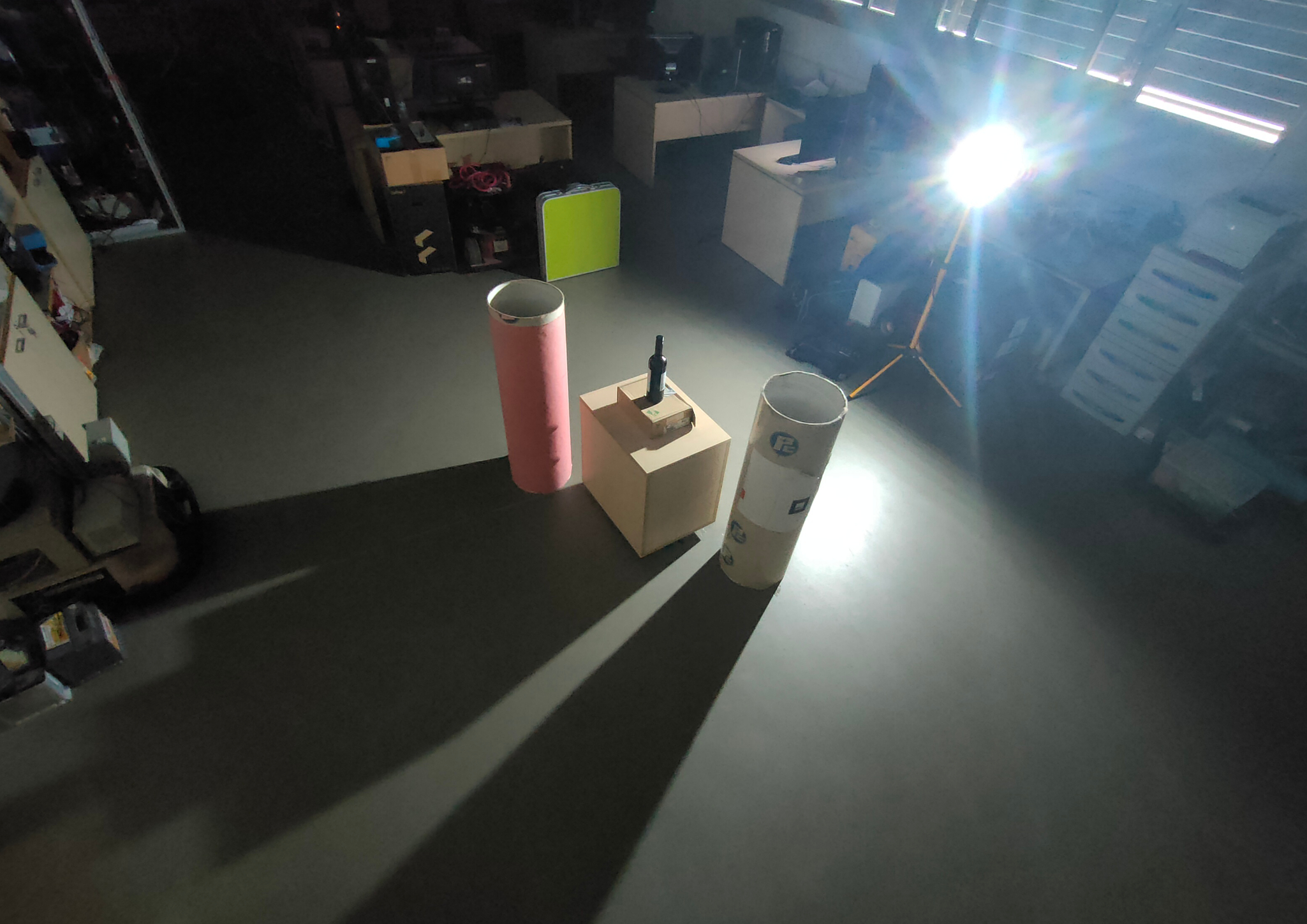}
     \end{subfigure}
      
    \caption{Real experiment setup. (Left) Ground robot platform. (Right) Working environment with occluders and a directional light. The robot moves to obtain measurements and decide whether there is an object at the target position.}
    \label{fig:real_experiment}
    \vspace{-20pt}
\end{figure}

\begin{figure}[t]
    \centering
    \begin{tabular}{cc}
        \includegraphics[width=0.45\columnwidth]{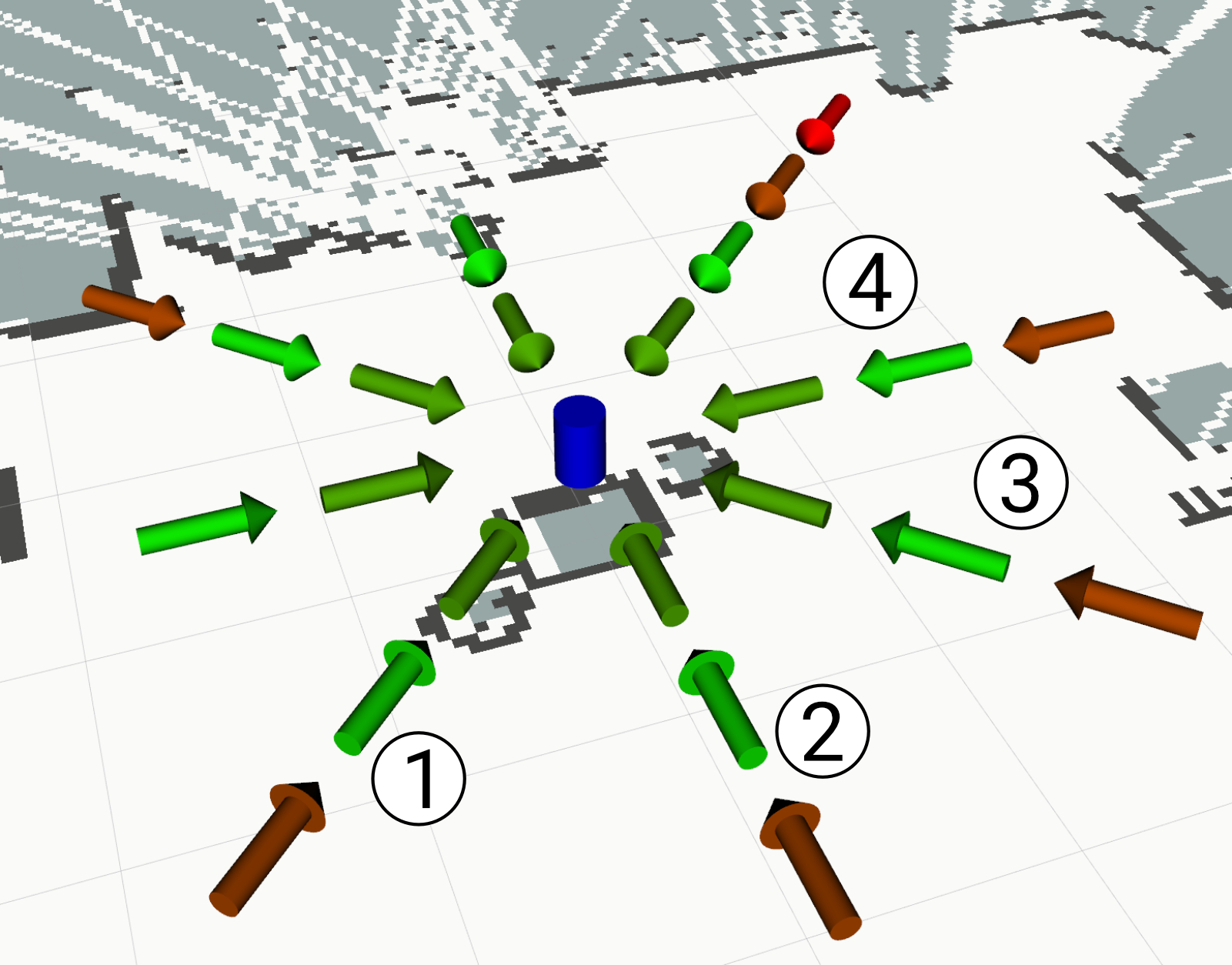} &
        \includegraphics[width=0.45\columnwidth]{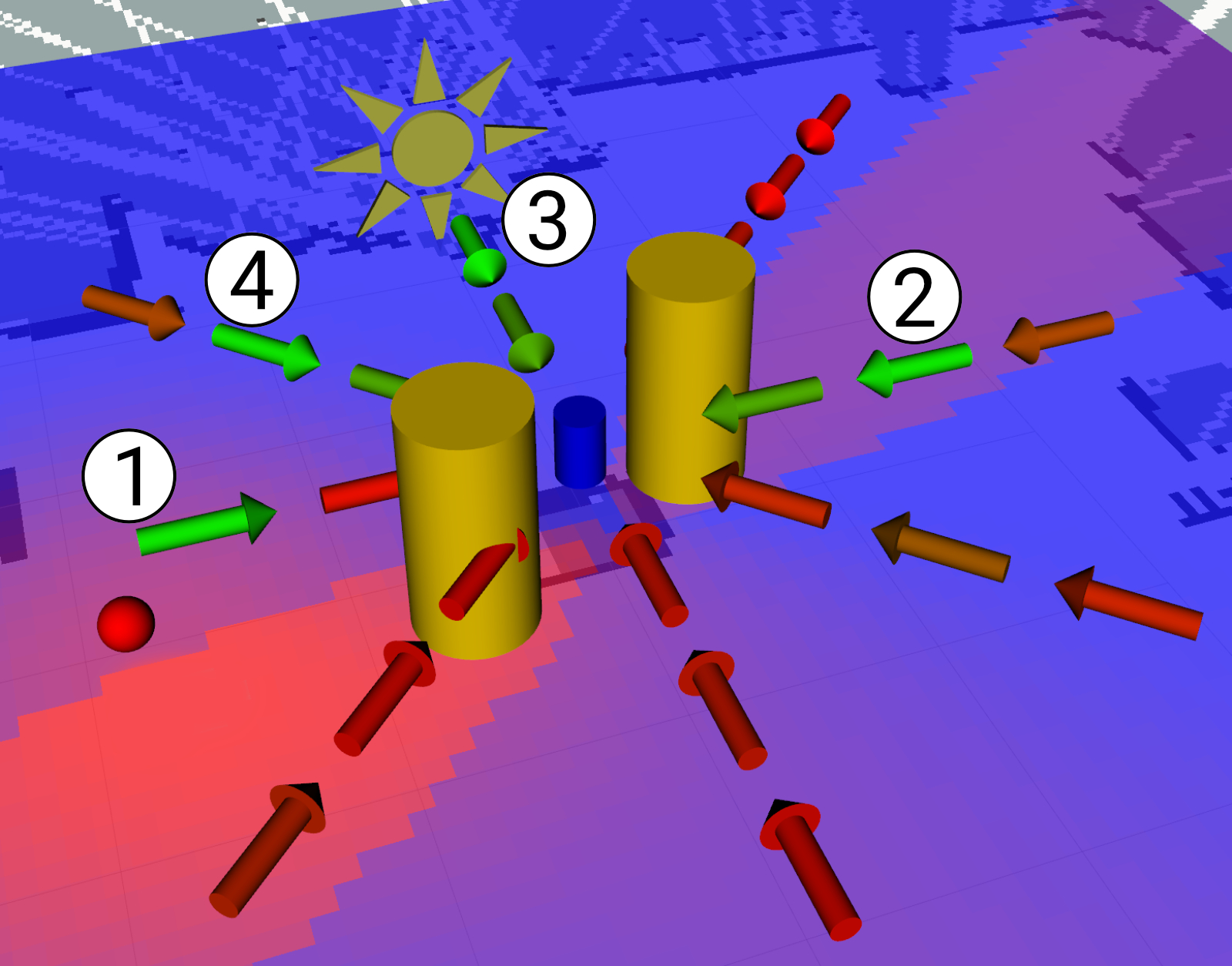} \\
        \includegraphics[width=0.45\columnwidth]{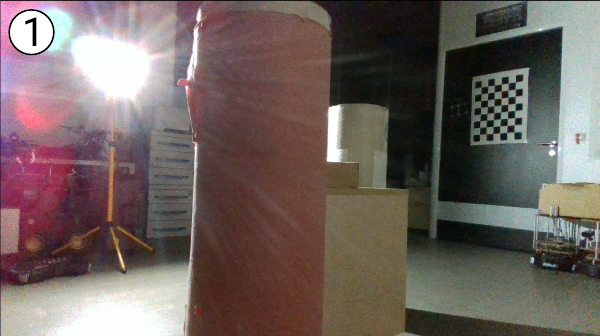} &
        \includegraphics[width=0.45\columnwidth]{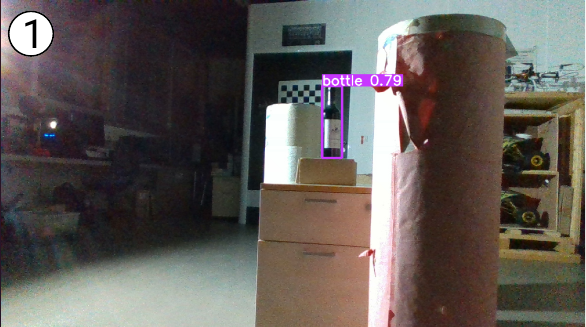} \\
        \includegraphics[width=0.45\columnwidth]{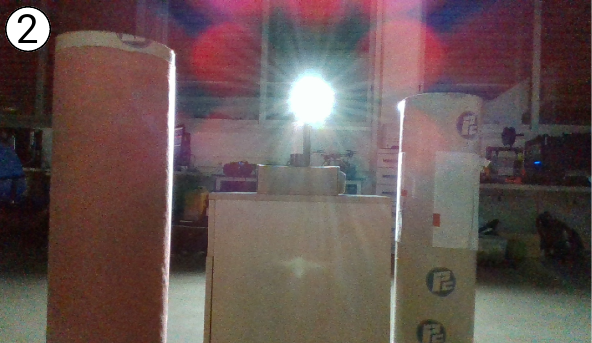} &
        \includegraphics[width=0.45\columnwidth]{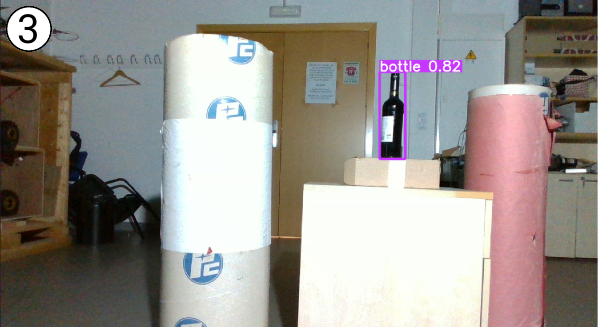} \\
    \end{tabular}
    \caption{Real experiment results. (Left) Basic IPP without perceptual factors. (Right) IPP considering perceptual factors for occluders (yellow cylinders), light sources (sun icon), and previous poses (red sphere), which contribute to the perceptual map projected on the floor. The arrows on the top figures show the candidate viewpoints to observe the blue target with colors indicating their estimated information value for the first step (green is better). The numbers indicate viewpoint selection order during the mission. The baseline leverages only a distance-based model for planning, while our approach includes perceptual information. As a result, the baseline obtains wrong measurements from the neural network (left 1 and 2) affected by occlusion (top) and back lighting (bottom). In contrast, our planner selects viewpoints that ensure good detections (right 1 and 3), resulting in better mission performance.}
    \label{fig:results_real}
    \vspace{-10pt}
\end{figure}

\subsection{Real-World Experiments}

The aim of this experiment is to showcase the applicability of our approach in real-world scenarios. \reffig{fig:real_experiment} shows our experimental setup. Our ground platform is a Turtlebot2 equipped with a Hokuyo 2D LiDAR to allow navigation in an occupancy map using the ROS \textit{navigation stack}. The robot is equipped with an Intel RealSene D435i RGB-D camera to perform object detection using the pre-trained YOLOv5 model from Ultralytics~\cite{yolov5}. All the processing is performed onboard the robot on an Intel NUC computer. The experiments are conducted in the room where there are two obstacles and a strong directional light.

The goal of the mission is to decide if there is a \textit{bottle} or \textit{no object} in a known position by taking multiple measurements. For that, the robot uses the active perception planner presented in~\refsec{SS:simulated_exp} to select viewpoints. Then, it navigates to them and captures an image. If the target object is detected by the neural network in the image, the detection confidence is integrated into the belief. Otherwise, a \textit{no object} observation is integrated with confidence $0.75.$ When the confidence for \textit{bottle} or \textit{no object} are $>0.99$, the mission ends. 
We compare the performance of two active planners, with and without all the perceptual factors described in the paper. Different from the simulation, candidate viewpoints are chosen from a deterministic set instead of using random sampling.

\reffig{fig:results_real} shows the result for the first viewpoint estimation values and the subsequent poses for the two planners. After four poses, due to the effect of occlusions and backlighting, the baseline only reaches an object detection confidence of $0.62$ and gets stuck moving between two viewpoints. In contrast, our planner obtains successful detections for all poses, integrating confidence over $0.99$. This shows the benefit of using our perceptual factors for informative viewpoint selection by considering correlations with the environment.
We refer the reader to the supplementary video for a complete visualization of the experiment.

\section{Conclusion and Future Work}
\label{sec:conclusion}

This paper presented a new perception model to characterize the impact that different environmental factors have on high-level measurements, like those provided by deep-learning vision algorithms.
To achieve this, we propose using perceptual factors, which are general functions used to quantify the aggregated influence of the environment on future measurements.
We demonstrated the integration of our perceptual factors into well-known recursive estimation filters and devised an active perception framework for informative viewpoint selection.
Further, we propose concrete mathematical examples of perceptual factors relevant for robotics scenarios.
We evaluated our complete active perception pipeline in simulation for robotic object localization and semantic classification tasks.
By incorporating perceptual factors, we obtain consistent state estimation and robust, informative viewpoint selection for predictive planning in active perception.
Potential avenues for future work include using deep learning to model perceptual factors and incorporating additional sensor modalities.

\balance

\bibliographystyle{style/IEEEtran} %
\bibliography{tex/references}
\end{document}